# SONIC: A Sparse Neural Network Inference Accelerator with Silicon Photonics for Energy-Efficient Deep Learning

Febin Sunny, Mahdi Nikdast, and Sudeep Pasricha
Colorado State University, Fort Collins, CO, USA
{febin.sunny, mahdi.nikdast, sudeep}@colostate.edu

*Abstract* – Sparse neural networks can greatly facilitate the deployment of neural networks on resource-constrained platforms as they offer compact model sizes while retaining inference accuracy. Because of the sparsity in parameter matrices, sparse neural networks can, in principle, be exploited in accelerator architectures for improved energy-efficiency and latency. However, to realize these improvements in practice, there is a need to explore sparsity-aware hardware-software co-design. In this paper, we propose a novel silicon photonics-based sparse neural network inference accelerator called SONIC. Our experimental analysis shows that SONIC can achieve up to 5.8× better performance-per-watt and 8.4× lower energy-per-bit than state-of-the-art sparse electronic neural network accelerators; and up to 13.8× better performance-per-watt and 27.6× lower energy-per-bit than the best known photonic neural network accelerators.

## I. Introduction

Over the past decade, convolutional neural networks (CNNs) have exhibited success in many application domains, such as image/video classification, object detection, and even sequence learning. As CNNs continue to be used for solving more complex problems, they have become increasingly compute and memory intensive. This is reflected in the increase in operations needed from ~4.5 million for LeNet-5 [1], proposed in 1998, to ~30 billion for VGG16 [2], proposed in 2014.

To keep pace with the continuous increase in CNN resource requirements, several accelerator platforms have been proposed, including graphical processing units (GPUs) with tensor units, Google's tensor processing units (TPUs), and custom application-specific integrated circuits (ASICs). However, these platforms still have low performance and energy-efficiency for most CNN applications. Sparse neural networks (SpNNs) [3] enable a reduced number of neurons and synapses while maintaining the original model accuracy. Therefore, they represent a promising optimization to reduce the overall resource requirements for CNNs in resource-constrained environments.

Unfortunately, simply deploying an SpNN on an accelerator does not necessarily ensure model performance and energy-efficiency improvements. This is because the strategies for dense neural network acceleration, for which most accelerators today are optimized, are not be able to take advantage of the sparsity available in neural networks. Dense neural network accelerators orchestrate dataflow and operations for parameters that have been sparsified (i.e., zeroed out). By having to process sparse parameters, conventional accelerators incur high latency and energy consumption that should be avoided. Therefore, carefully devised strategies for taking advantage of sparsity and reducing the number of operations becomes absolutely essential.

There have been a few recent efforts to design accelerators that provide support for SpNNs [4]-[6]. But these electronic accelerators face fundamental limitations in the post-Moore era, where processing capabilities are no longer improving as they once did, and metallic wires create new dataflow bottlenecks [7]. Neural network accelerator architectures that leverage silicon photonics for computing and data transfer can enable low latency and energy-efficient computation solutions [8]-[10]. However, they are not impervious to the high latency and energy wastage problem when accelerating SpNNs.

In this work, we present a novel neural network accelerator designed with silicon photonics that is optimized for exploiting sparsity, to enable energy-efficient and low-latency SpNN acceleration. To the best of our knowledge, this work presents the first non-coherent photonic SpNN accelerator (see Section II). Our novel contributions in this paper include:

- The design of a novel photonics-domain SpNN hardware accelerator architecture that utilizes a modular, vector-granularity-aware structure to enable high throughput and energy-efficient execution across different CNN models;
- Sparsity-aware data compression and dataflow techniques for fully connected and convolution layers, which are tuned for the high throughput operation of our photonic accelerator;
- A comprehensive comparison with state-of-the-art sparse electronic and dense photonic CNN accelerator platforms, to demonstrate the potential of our accelerator platform.

## II. Related Work

To efficiently accelerate SpNNs, there is a need for specialized hardware architectures. In recent years, a few such architectures have been proposed by the electronic machine learning (ML) acceleration research community, e.g., [4]-[6]. The framework presented in [4] leveraged a custom instruction-set architecture (ISA) for SpNNs. Specialized buffer controller architectures were used, which involved indexing to keep track of sparse elements, thus preventing them from being fed to the processing elements. In [5], a software-hardware co-optimized reconfigurable sparse CNN accelerator design was proposed for FPGAs. The architecture exploited both inter- and intra-output feature map parallelism. Kernel merging along with structured sparsity were considered to further improve overall efficiency. The work in [6] described an FPGA-based implementation of a sparse CNN accelerator. The accelerator made use of an output feature map compression algorithm, which allowed the accelerator to operate directly on compressed data.

To obtain lower latency and better energy efficiency, there has been growing interest in using silicon photonics for ML

This research is supported by grants from NSF (CCF-1813370, CCF-2006788)

acceleration. Silicon photonic neural network accelerators can be broadly classified into two types: *coherent* and *non-coherent*. Coherent architectures use a single wavelength to operate and imprint weight and activation parameters onto the electrical field amplitude of optical signals, e.g., [9]. In contrast, non-coherent architectures use multiple wavelengths, where each wavelength can be used to perform an individual neuron operation in parallel with other wavelengths, e.g., [8], [10]. In these architectures, parameters are imprinted directly onto signal amplitude. The recent work in [9] was the first to consider sparsity in the design of coherent photonic neural network accelerators. Structured sparsity techniques were used along with fast Fourier transform (FFT) based optical convolution, with the goal of reducing the area consumption of coherent architectures. The singular value decomposition (SVD)-based approach for phase matrix representation, which is crucial to reduce the overall area of the coherent architecture, also makes these architectures susceptible to accuracy loss, as the experiments in [9] showed. Due to phase encoding noise, phase error accumulation, and scalability limitations of coherent accelerators [24], there has been a growing interest in non-coherent photonic accelerators. Non-coherent dense neural network accelerators were proposed in [8] and [10], where the basic device for multiply and accumulate units relies on microring resonators [8] and microdisks [10]. The work in [8] also utilized cross-layer device-level and circuit-level optimizations to enable lower power consumption in the optical domain. *The SONIC architecture proposed in this work represents the first non-coherent photonic SpNN accelerator, where multiple software optimizations for sparsity, clustering, and dataflow are integrated closely with the hardware architecture design for improved energy-efficiency and latency, without compromising on inference accuracy.*

The rest of the paper is organized as follows. Section III provides an overview of our proposed software and dataflow optimizations for convolution and fully connected layers in CNNs. Section IV describes the hardware design of our non-coherent photonic accelerator that is tuned for these model optimizations. Section V presents the experiments conducted and results. Lastly, we draw conclusions in Section VI.

## III. Software and Dataflow Optimizations

### A. Model Sparsification

To generate SpNNs, we adapt a layer-wise, sparsity-aware training approach from [11]. We opt for layer-wise sparsity instead of sparsifying the entire model, to have more control over the process, and to avoid overly sparsifying sensitive layers which notably contribute to the overall model accuracy. In our approach, for every layer selected to be sparsified, a binary mask variable is added, which is of the same size and shape as the layer's weight tensor. The algorithm also determines which of the weights participate in the forward execution of the graph. The weights in the chosen layer are then sorted by their absolute values and the smallest magnitude weights are masked to zero until the user-specified sparsity levels are reached. Also note that we opt for sparsity-aware training instead of post-training sparsification, as the latter approach can indiscriminately remove neurons, thus adversely affecting the inference accuracy. We also utilize an L2 regularization term during training, to encourage smaller weight values and avoid overfitting, which further helps improve the overall accuracy post-deployment.

### B. Weight Clustering

The electrical-optical interface in photonic accelerators, such as in [8], can be highly power consuming. This is because digital-to-analog converters (DACs), which have high power overheads, are used in these interfaces to tune the optical devices in the multiply-and-accumulate (MAC) units. Moreover, higher resolution (i.e., the number of bits used to represent each weight and activation parameter) requirements for a DAC translate into higher power and latency overheads in the DAC. Thus, to reduce these DAC overheads, we perform post-training quantization of the models, in the form of weight clustering. We opt for density-based centroid initialization of the weights, for the clustering operation, as described in [12]. For this clustering approach, a cumulative distribution function is built for the weights. The distribution is evenly divided into regions, based on the user specified number of clusters. The centroid weight values of the evenly distributed regions are then deduced, and these values are used to initialize clustering. This process effectively reduces the variations in weight values and confines the values to the centroids. Therefore, if there are $C$ centroids, and thus $C$ clusters, the model will end up with $C$ unique weights. This implies that the weights can be represented with a resolution of $\log_2 C$, thus reducing the required DAC resolution and enabling power and latency savings. Section V.A describes our weight clustering (and sparsification) explorations and parameters in more detail.

### C. Dataflow Optimizations

Beyond sparsification and weight clustering optimizations, we also perform enhancements to improve dataflow efficiency in our hardware platform. Fully Connected (FC) layers are computationally intensive layers in CNNs where all neurons in the layer are connected to all the other neurons in the following FC layer. The baseline FC-layer operation is a matrix-vector product, which generates the output vector to be passed on to the next FC layer, as represented in Fig. 1(a). As the figure shows, there can be many parameters in the weight matrix and the activation vector which are zeroes. These zero parameters can be prevented from being passed on to the processing elements to reduce model latency and energy consumption.

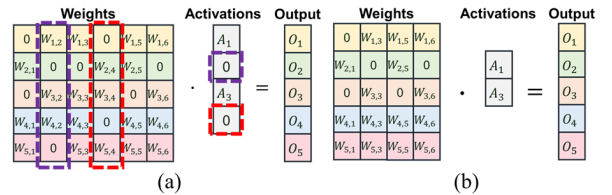

Fig. 1: FC layer operation, where the product of the weight matrix and activation vector is calculated. (a) Zero element identification in the activation vector and corresponding columns in weight matrix (marked in dotted outlines); (b) Compressed matrix and vector, but weight matrix still exhibits parameter sparsity.

To achieve this goal, a compression approach as depicted in Figs. 1(a)-(b) is utilized. In this approach, we identify the zero parameters in the activation vector, and remove the corresponding columns in the weight matrix which will be operated upon by these parameters during the dot-product operation. This approach generates dense activation vectors, but the weight vectors can still be sparse, as depicted in the weight matrix in Fig. 1(b). This process also does not impact the output vector calculation accuracy or output vector dimension.

For convolution (CONV) layers, the main difference with FC layers is the convolution operation performed in CONV layers. We unroll the CONV layer kernels and their associated patch of the input feature (IF) map matrix, to form vector-dot-product operations from the convolution operations (see Fig. 2(a)). The compression approach for FC layers can be repeated for these unrolled matrix-vector multiplication operations (see Fig. 2(b)). The compression approach in CONV layers helps generate dense kernel vectors to be passed to the vector-dot-product units (VDUs). Note that the IF vectors (activations) being passed for processing may still have sparsity present, as shown in Fig. 2(c).

The residual sparsity in the FC layer weight matrices and the CONV layer IF maps is handled at the vector-dot-product unit (VDU) level, as discussed in more detail in Section IV.B.

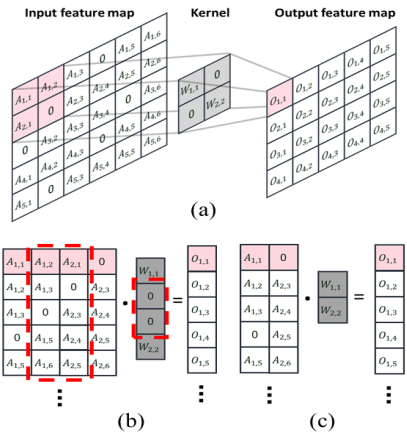

Fig. 2: (a) Convolution operation between kernel (weight) matrix and input feature map (activations). A patch of the input feature map is convolved with the kernel matrix at a time to generate an output feature map element (a patch and the corresponding output element are shown in red boxes). (b) Convolution operation unfurled into a vector-matrix-dot-product operation; avenues for compression are indicated by dotted-red outlines. (c) The result of the compression approach, with input feature map still exhibiting parameter sparsity.

## IV. *SONIC* Hardware Accelerator Overview

Fig. 3 shows a high-level overview of the proposed non-coherent *SONIC* architecture for SpNN inference acceleration. *SONIC* comprises of an optical processing core, which uses vector-dot-product units (VDUs)—described in Section IV.B—to perform multiply and accumulate operations for FC and CONV layers in the photonic domain during inference. Several peripheral electronic modules are also integrated, to interface with the main memory, map the dense and sparse vectors to the photonic VDUs, and perform post-processing operations, such as applying non-linearities and accumulating partial sums generated by the photonic core. DAC arrays within VDUs convert buffered signals into analog tuning signals for MRs, and vertical-cavity surface-emitting lasers (VCSELs) are used to generate different wavelengths. Analog-to-digital converter (ADC) arrays are used to map the output analog signals generated by photonic summation to digital values that are sent back for post-processing and buffering. The devices, VDU, and architecture are discussed further in the following subsections.

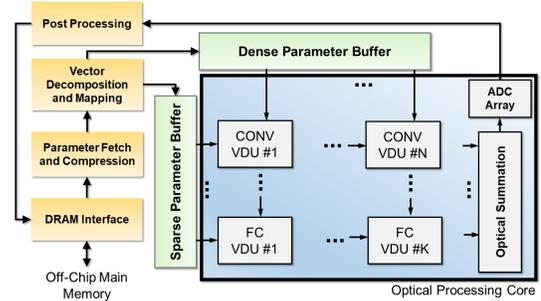

Fig. 3: An overview of the *SONIC* architecture, with *N* CONV layer-specific VDUs and *K* FC layer-specific VDUs.

### A. Microring Resonators (MRs) and Robust Tuning

MRs are the primary devices used within our VDUs to implement matrix-vector multiplication operations. MRs are wavelength-selective silicon photonic devices, which are usually designed to be responsive to a specific 'resonant' wavelength ($\lambda_{MR}$). Such MRs are used to modulate and filter their resonant wavelengths in a carefully controlled manner, via a tuning circuit, to realize multiplications in the optical domain.

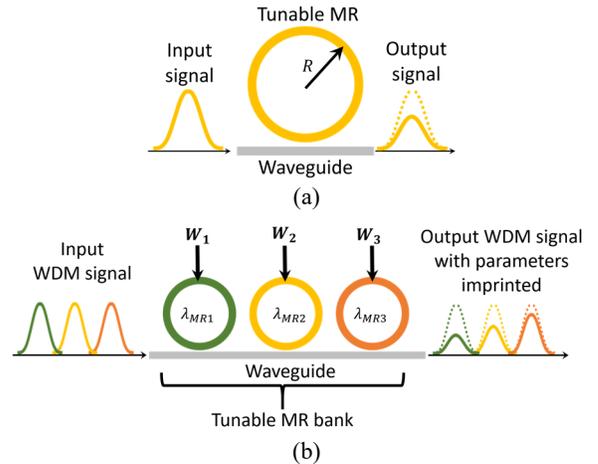

Fig. 4: (a) An all-pass microring resonator (MR) filter ($R$ is the radius of the MR and determines the resonant wavelength). (b) An MR bank where multiple MR filters, each sensitive to a particular wavelength, are arranged to perform vector-matrix multiplication.

An MR tuning mechanism can be used to induce a resonance shift ($\Delta\lambda_{MR}$), and to change the output wavelength amplitude (Fig. 4(a)) to realize a scalar multiplication operation. The tuning mechanism in MRs operates by heating (thermo-optic (TO) tuning [13]) or carrier injection (electro-optic (EO) tuning [14]), thereby inducing a change in effective index ($n_{eff}$), which impacts $\lambda_{MR}$. The induced $\Delta\lambda_{MR}$ increases the loss a wavelength experiences as it passes the MR, modifying the amplitude and imprinting the desired parameter ($W_1$–$W_3$ for $MR_1$–$MR_3$ in Fig 4(b)). To improve throughput, WDM signals are used with a group of MRs (i.e., MR bank, Fig. 4(b)), where each MR is sensitive to a specific $\lambda_{MR}$. A large passband in MRs can be achieved by cascading several of them, as in [15], which can be used to simultaneously tune multiple wavelengths.

In *SONIC,* we make use of a hybrid tuning circuit where both TO and EO tuning are used to induce $\Delta\lambda_{MR}$. Such a tuning approach has previously been proposed in [16] for silicon photonic devices with low insertion loss. This approach can be easily transferred to MR banks for hybrid tuning in our architecture. The hybrid tuning approach supports faster operation of MRs with fast EO tuning to induce small $\Delta\lambda_{MR}$ and using TO tuning for large $\Delta\lambda_{MR}$. To further reduce the power overhead of TO tuning in our hybrid approach, we adapt a method called thermal eigen decomposition (TED), which was first proposed in [17]. Using TED, we can collectively tune all the MRs in an MR bank with much lower power consumption.

*B. Vector-Dot-Product Unit (VDU) Design*

As we decompose the operations in FC and CONV layers to vector-dot-product operations, our processing units are effectively vector-dot-product units (VDUs). Fig. 5 depicts the VDU design in *SONIC*. As Figs. 1(b) and 2(c) showed, the granularity of the vectors involved in FC and CONV operations can be different. In real models, the CONV kernel sizes are relatively small when compared to FC layers. Also, in our dataflow for CONV layers, the dense vectors are generated by kernel matrices (weights), and for FC layers, the dense vectors are generated by activation vectors. However, for CONV layer dense vectors, we only need low resolution digital-to-analog converters (DACs), because of the clustering approach we utilize (see Section III.B). Moreover, for FC layers, the sparse vectors may utilize the low-resolution DACs, due to the same reason. Therefore, considering these differences, we separate the VDU implementations for CONV and FC layers. However, both the VDU implementations follow the layout illustrated in Fig. 5.

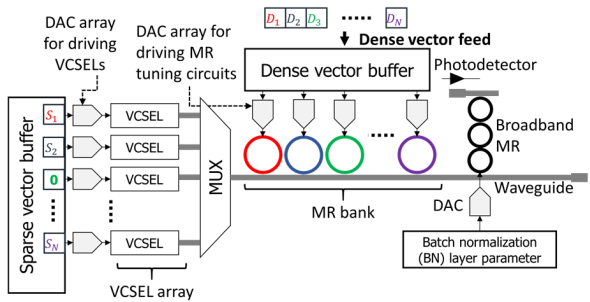

Fig. 5: Vector-dot-product unit in the *SONIC* architecture.

As shown in Fig. 5, VDUs use separate local buffers to store the sparse and dense vector parameter values. The parameters are fed into DAC arrays for driving the optical devices (MRs or VCSELs). Each VDU has a local VCSEL array, which is driven using a DAC array. A DAC drives its corresponding VCSEL to generate optical signals with amplitude tuned to reflect its corresponding vector parameter.

We enhance the VDU design for sparsity by preventing a VCSEL from being driven if a zero element is encountered in the sparse vector (recall that after the compression approach described in Section III.C, there may be residual sparsity in the IF map or weight matrix). This involves power gating the VCSEL, and hence subsequent operations for the dot product will not occur. The power gating thus helps avoid the wasteful operations with the zero parameters, which were not eliminated by our data compression approach (see Section III), in the vectors fed to the VDU. The signals from the VCSEL array are fed into an optical multiplexer (MUX in Fig. 5) to generate a WDM signal, which is transmitted to the MR bank via a waveguide. The MR bank is comprised of several tunable MRs, each of which can be tuned to alter the optical signal amplitude of a specific input wavelength, so that the intensity of the wavelength reflects a specific value, as discussed earlier.

We also make use of a broadband MR to tune all wavelengths simultaneously to reflect batch normalization (BN) parameters for a layer (Fig. 5). Once the multiplication between parameters and BN parameters have been performed, a photodetector is used to convert the optical signal back to an electrical signal, to obtain a single, accumulated value from the VDU.

*C. SONIC Architecture*

The VDU design discussed above is integrated in the *SONIC* architecture shown in Fig. 3. As mentioned earlier, we separate the VDU designs for the FC and CONV layer operations. The separate VDU designs account for the vector granularity differences between FC and CONV layer operations, and the differences in DAC requirement for driving the VCSEL and MR arrays. The architecture relies on an electronic-control unit for interfacing with the main memory, retrieving the parameters, mapping the compressed parameters, and post-processing the partial sums generated by the VDUs. The optical processing core (see Fig. 3) focuses on CONV, FC, and batch normalization acceleration during the inference phase. Other operations, such as activation and pooling are implemented electronically, as done in all prior works on optical computation, due to the difficulty in performing such operations optically. The *SONIC* architecture design in Fig. 3 arranges VDUs in an array. For CONV layers, we consider $N$ VDU units, with each unit supporting an $n \times n$ dot product. For FC layer acceleration, we consider $K$ VDU units, with each unit supporting a $m \times m$ dot product. Here, $m > n$ and $N > K$, as per the requirements of each of the distinct layers. In each VDP unit, the original vector dimensions are decomposed into $n$ or $m$ dimensional vectors. Here, $n$ and $m$ are dependent on the dense vector granularity we obtain through the compression approach for the CONV and FC layers, as discussed in Section III.C.

## V. Experiments and Results

For our experiments, we consider four custom CNN models with both CONV and FC layers, for the well-known CIFAR10, STL10, SVHN, and MNIST datasets. Details on the baseline models are shown in Table 1. For evaluating the performance of the *SONIC* architecture, we compare it against two state-of-the-art SpNN accelerators: *RSNN* [5] and *NullHop* [6], along with dense photonic accelerators *CrossLight* [8] and *HolyLight* [10], and a photonic binary neural network accelerator *LightBulb* [23]. Furthermore, we attempted to implement the coherent SpNN photonic accelerator from [9]; however, the work does not provide details or results for latency, power, and energy, which prevented us from comparing against it. We also show comparative results against the NVIDIA Tesla P100 GPU and Intel Xeon Platinum 9282 CPU. We compared all these architectures in terms of throughput (i.e., frame per second (FPS)), energy per bit (EPB), and power consumption efficiency (FPS/W). We devised a custom Python simulator, integrated with Tensorflow v2.5, to evaluate *SONIC* and other accelerators. The parameters summarized in Table 2 were used to configure the accelerators to obtain performance and power/energy results.

Table 1: CNN models considered for experiments.

| Datasets | Conv layers | FC layers | No. of parameters | Baseline accuracy |
|---|---|---|---|---|
| MNIST | 2 | 2 | 1,498,730 | 93.2% |
| CIFAR10 | 6 | 1 | 552,874 | 86.05% |
| STL10 | 6 | 1 | 77,787,738 | 74.6% |
| SVHN | 4 | 3 | 552,362 | 94.6% |

Table 2: Parameters considered for analysis of accelerators.

| Devices | Latency | Power |
|---|---|---|
| EO Tuning [13] | 20 ns | 4 $\mu$W/nm |
| TO Tuning [14] | 4 $\mu s$ | 27.5 mW/FSR |
| VCSEL [18] | 0.07 ns | 1.3 mW |
| Photodetector [19] | 5.8 ps | 2.8 mW |
| DAC (16 bit) [20] | 0.33 ns | 40 mW |
| DAC (6 bit) [21] | 0.25 ns | 3 mW |
| ADC (16 bit) [22] | 14 ns | 62 mW |

*A. Model Sparsification and Clustering Results*

In the first experiment, we focus on software model optimization in *SONIC*. To obtain the best accuracy possible, we performed layer-wise sparsification in the models considered, as described in Section III.A. We also use this experiment to partially explore the design space of *SONIC* hardware implementations. As depicted in Fig. 5, we use DACs for driving MRs and VCSELs in our accelerator. To decide on the required DAC resolution (and corresponding power and latency costs), we perform post-training weight clustering, as described in Section III.B. Our goal was to generate models with as much per-layer sparsity as possible, and minimal DAC resolution, while exhibiting comparable accuracy to the baseline model.

Table 3: Summary of the sparsification and clustering results.

| Datasets | Layers pruned | No. of weight clusters | No. of parameters | Final accuracy |
|---|---|---|---|---|
| MNIST | 4 | 64 | 749,365 | 92.89% |
| CIFAR10 | 7 | 16 | 276,437 | 86.86% |
| STL10 | 5 | 64 | 46,672,643 | 75.2% |
| SVHN | 5 | 64 | 331,417 | 95% |

A summary of the optimized models and the final accuracy achieved after sparsification and weight clustering is shown in Table 3. Note that the final accuracy of the optimized models in Table 3 is comparable or slightly better than the baseline accuracy shown in Table 1, which is consistent with the trend in prior works. To arrive at these numbers for each model, we performed a detailed exploration. Fig. 6 shows the design space considered during sparsity and clustering exploration for the CIFAR10 model (figures for the other three models are omitted for brevity). The best solution selected is the one with the highest accuracy and is highlighted with a star (the same solution as shown in Table 3). Fig. 7 further shows the layer-wise breakdown of sparsification for all four models, where the plots show the layer-specific sparsity level for weight parameters (in the best solution for each model from Table 3) and the resulting sparsity in activations as they traverse the sparse layers. Our exploration was able to identify the need for a maximum of 16 clusters for the best CIFAR10 solution and a maximum of 64 clusters across the four models (Table 3). Based on these results, we consider 6-bit DACs (to support up to 64 levels) for weight parameters. We kept activation granularity at 16-bits, which provided us with sufficient accuracy (Table 3) and thus used 16-bit DACs for activations, in the *SONIC* accelerator.

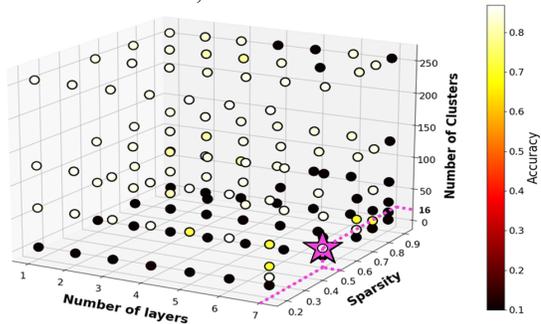

Fig. 6: Visualization of sparsity and clustering exploration on the CIFAR10 model. Number of layers is the total layers sparsified, sparsity is the average pruning aggressiveness, and number of clusters refers to the total weights clusters. The best (highest accuracy) configuration is indicated by the star.

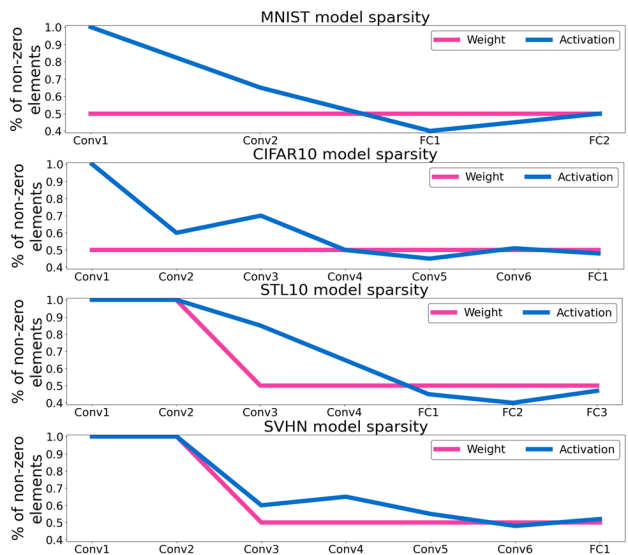

Fig. 7: Sparsity across various layers in the four models considered.

*B. Comparison with state-of-the-art accelerators*

We explored various *(n, m, N, K)* configurations for the *SONIC* architecture (see Section IV.C) and found the best configuration in terms of FPS/W, EPB, and power consumption to be (5, 50, 50, 10). We found that the value of *n* is heavily dependent on CONV layer kernel values, which was fixed after our model sparsification experiments. Increasing *n* beyond five did not provide any benefits, as the dense kernel vectors do not exceed five-parameter granularity for the considered models.

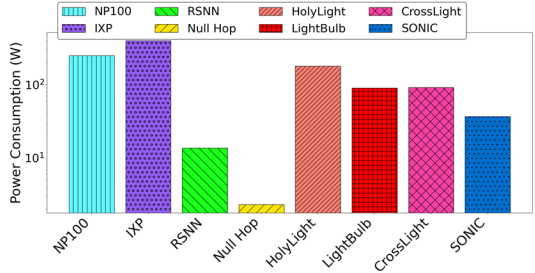
Fig. 8: Power comparison across the accelerator platforms.

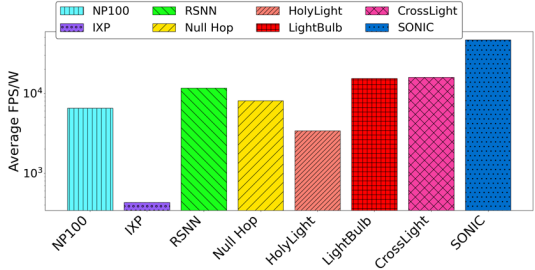
Fig. 9: FPS/W comparison across the accelerator platforms.

Fig. 8 shows power consumption and Fig. 9 shows the power efficiency (in terms of frames-per-second/watt or FPS/W) across the accelerators considered. In these figures, NP100 is the GPU and IXP is the CPU. We can observe that due to its sparsity-aware, clustering-aware, and dataflow-optimized hardware architecture design, *SONIC* exhibits substantially higher power efficiency, even though it has higher power consumption than the electronic SpNN accelerators. *SONIC*, on average, exhibits 5.81× and 4.02× better FPS/W than the *NullHop* and *RSNN* electronic SpNN accelerators. *SONIC* also exhibits 3.08×, 2.94×, and 13.8× better power efficiency on average than the *LightBulb*, *CrossLight*, and *HolyLight* photonic accelerators, respectively. This is because none of these photonic accelerators are optimized to take advantage of sparsity and clustering.

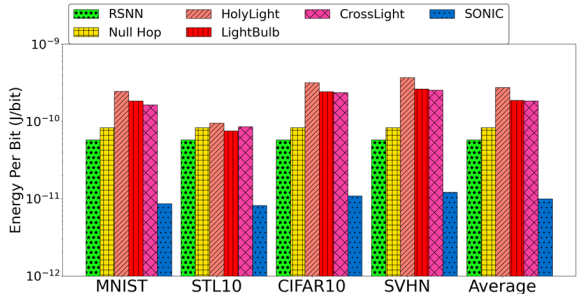
Fig. 10: EPB comparison across the accelerator platforms.

When comparing the energy-per-bit (EPB) across the accelerators, as shown in Fig. 10, we can again observe that the co-design of the software and dataflow optimizations along with the hardware architecture in *SONIC* allow it to outperform the photonic and electronic SpNN accelerators. *SONIC* exhibits, on average, 19.4×, 18.4×, and 27.6× lower EPB than *LightBulb*, *CrossLight*, and *HolyLight*, respectively. *SONIC* also exhibits 8.4× and 5.78× lower EPB than *NullHop* and *RSNN*. These results highlight the promise of *SONIC* for optimized SpNN implementations on resource-constrained platforms.

## VI. Conclusions

In this paper, we presented a novel non-coherent photonic sparse neural network accelerator, called *SONIC*, that integrates several hardware and software optimizations. *SONIC* exhibits up to 5.8× better power efficiency, and 8.4× lower EPB than state-of-the-art sparse electronic neural network accelerators; and up to 13.8× better power efficiency and 27.6× lower EPB than state-of-the-art dense photonic neural network accelerators. These results demonstrate the promising low-energy and low-latency inference acceleration capabilities of our *SONIC* architecture.